\documentclass{article}
\usepackage{ijcai24}

\usepackage{times}
\usepackage{soul}
\usepackage{url}
\usepackage[hidelinks]{hyperref}
\usepackage[utf8]{inputenc}
\usepackage[small]{caption}
\usepackage{graphicx}
\usepackage{amsmath}
\usepackage{amsthm}
\usepackage{booktabs}
\usepackage{algorithm}
\usepackage{algorithmic}
\usepackage[switch]{lineno}
\usepackage[T1]{fontenc}
\usepackage{amsmath}
\usepackage{color}
\usepackage{enumitem}
\usepackage{times}
\usepackage{latexsym}
\usepackage{multirow, multicol}
\usepackage{microtype}
\usepackage{graphicx}
\usepackage{bbding}
\usepackage{inconsolata}
\usepackage{tabularx}
\usepackage{booktabs}
\usepackage{makecell}

\title{SpeechComposer:\\ Unifying Multiple Speech Tasks with Prompt Composition}

\author{
Yihan Wu$^{2,1}$\thanks{Work done during the visit at CMU}\and
Soumi Maiti$^1$\and
Yifan Peng$^1$\and
Wangyou Zhang$^{3,1}$\and
Chenda Li$^{3,1}$\and
Yuyue Wang$^2$\and
Xihua Wang$^2$\and
Shinji Watanabe$^1$\and
Ruihua Song$^2$\\
\affiliations
$^1$Carnegie Mellon University, USA \\
$^2$Renmin University of China, China \\
$^3$Shanghai Jiao Tong University, China \\
}

\begin{document}

\maketitle

\begin{abstract}

Recent advancements in language models have significantly enhanced performance in multiple speech-related tasks. Existing speech language models typically utilize task-dependent prompt tokens to unify various speech tasks in a single model. However, this design omits the intrinsic connections between different speech tasks, which can potentially boost the performance of each task. In this work, we propose a novel decoder-only speech language model, SpeechComposer, that can unify common speech tasks by \textit{composing a fixed set of prompt tokens}. Built upon four primary tasks --- speech synthesis, speech recognition, speech language modeling, and text language modeling --- SpeechComposer can easily extend to more speech tasks via compositions of well-designed prompt tokens, like voice conversion and speech enhancement. The unification of prompt tokens also makes it possible for knowledge sharing among different speech tasks in a more structured manner. Experimental results demonstrate that our proposed SpeechComposer can improve the performance of both primary tasks and composite tasks, showing the effectiveness of the shared prompt tokens. Remarkably, the unified decoder-only model achieves a comparable and even better performance than the baselines which are expert models designed for single tasks.

\end{abstract}

\section{Introduction}
Recently, large language models show remarkable performance across diverse domains, including natural language processing~\cite{OPT,GPT32023Tom,llama2023Touvron}, vision~\cite{singh2022flava}, and multi-modal tasks~\cite{blip22023li,alayrac2022flamingo}. Predominantly employing an autoregressive manner and the decoder-only architecture, these models transform various tasks into generative ones. Despite the simplicity of their training methodologies, they exhibit exceptional capabilities.

\begin{figure}[t]
    \centering
    \includegraphics[width=\linewidth]{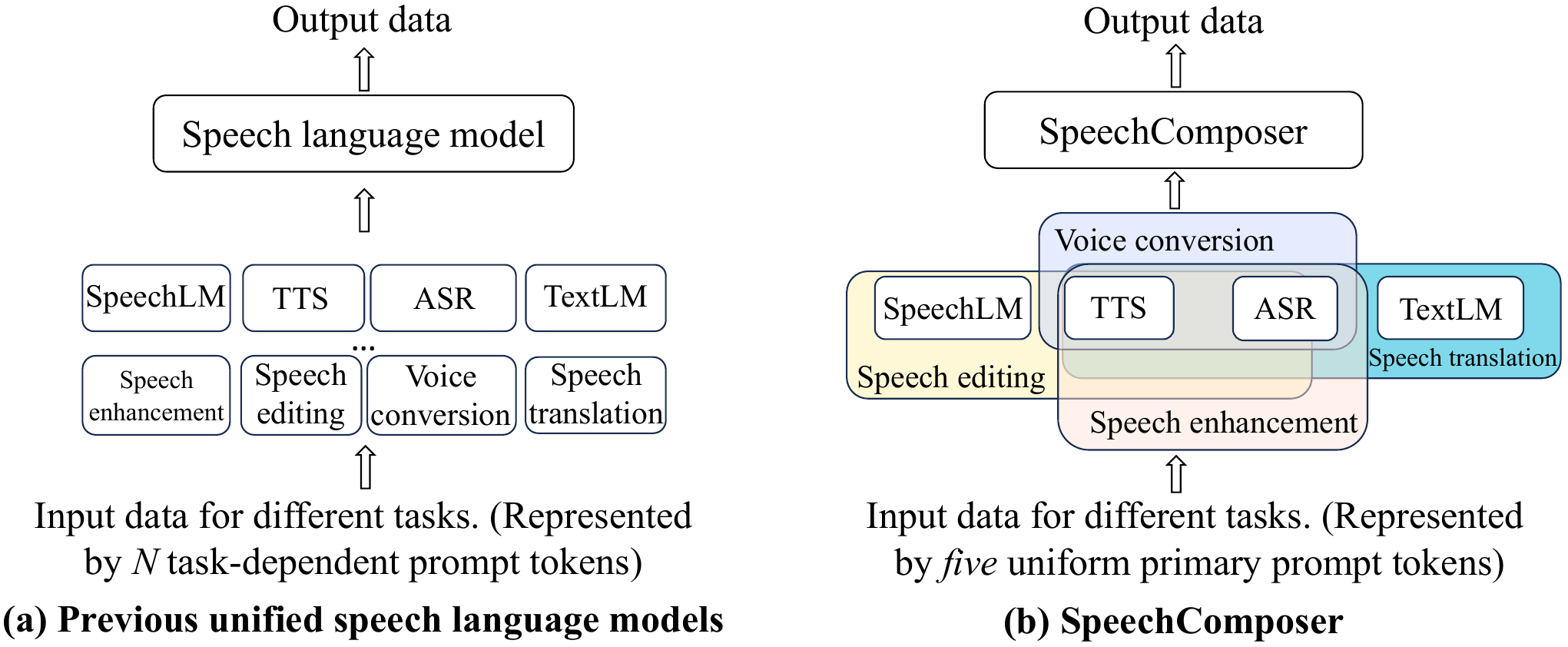}

    \caption{The comparison of SpeechComposer with other multi-task speech models when performing $N$ tasks. (a) shows a cascaded model that requires different sub-models for each task. (b) identifies previous works using $N$ task-dependent prompt tokens for each task. (c) shows that our proposed SpeechComposer uses a unified architecture and composes primary tasks to more tasks with only five uniform prompt tokens.}
    \label{fig:related_works}
    \vspace{-1 em}
\end{figure}

\begin{table*}[t]
\centering
\resizebox{\linewidth}{!}{\begin{tabular}{c|c|c|c|c}
\toprule
 &\textbf{Task} & \textbf{Task composition} & \textbf{Data format} & \textbf{Evaluation metrics}  \\
\midrule
\multirow{4}{*}{Primary tasks} & TextLM & ---  & \texttt{\textlangle generate-text\textrangle}, $Y$ & PPL  \\
& SpeechLM & ---  & \texttt{\textlangle generate-speech\textrangle}, $D$ & PPL \\
& ASR & ---  & \texttt{\textlangle start-speech\textrangle}, $D$, \texttt{\textlangle generate-text\textrangle}, $Y$ & WER \\ 
& TTS & ---  & \makecell{\texttt{\textlangle start-text\textrangle}, $Y$, \textcolor{cyan}{\texttt{\textlangle enroll-speech\textrangle}}, \textcolor{cyan}{$D^{\text{enroll}}$}, \\ \texttt{\textlangle generate-speech\textrangle}, $D^{\text{tgt}}$ } & CER, MOSNet  \\
\midrule
\midrule
\multirow{2}{*}{Composite tasks} & \makecell{VC} & ASR $+$ TTS & \makecell{\texttt{\textlangle start-speech\textrangle}, $D^{\text{src}}$, \texttt{\textlangle generate-text\textrangle}, $Y$, \\ \textcolor{cyan}{\texttt{\textlangle enroll-speech\textrangle}}, \textcolor{cyan}{$D^{\text{enroll}}$}, \texttt{\textlangle generate-speech\textrangle}, $D^{\text{tgt}}$} & \makecell{MOSNet, CER,\\Sim-\textit{speech}, TER} \\
\cline{2-5}
& \makecell{SE} & ASR$+$TTS &  \makecell{\texttt{\textlangle start-speech\textrangle}, $D^{\text{noise}}$, \texttt{\textlangle generate-text\textrangle}, $Y$, \\ \textcolor{cyan}{\texttt{\textlangle enroll-speech\textrangle}}, \textcolor{cyan}{$D^{\text{enroll}}$}, \texttt{\textlangle generate-speech\textrangle}, $D^{\text{clean}}$} & DNSMOS, WER \\
\bottomrule
\end{tabular}
}
\caption{SpeechComposer's data format for different tasks (primary tasks and composite tasks) in the training stage and inference stage. Here, $Y$ and $D$ refer to the input text sequence and discrete speech tokens separately. The input format for speech enhancement and voice conversion is the same, but manipulating different enrollment speech $D^{\text{enroll}}$ to control the generated speech. For the speech enhancement task, $D^{\text{enroll}}$ is a clean speech sample from the same speaker during training, while in voice conversion, $D^{\text{enroll}}$ is a speech utterance from the target speaker.}

\label{tab:data_format}
\end{table*}

The current works on speech language models mainly treat speech tasks as conditional generation tasks. It involves encoding the speech signal into a discrete
representation~\cite{Alexei2020wav2vec2,hsu2021hubert} and modeling it with language models~\cite{Valle2023wang,SpeechX2023wang,makeavoice2023huang,Audiolm2023Zal,GSLM2021lakhotia}.
Taking advantage of large-scale speech data, some recent studies~\cite{SpeechX2023wang,Vallex2023zhang,polyvoice2023dong} extend the application of speech language models to multiple tasks, integrating multi-task learning into a unified model (as shown in Figure~\ref{fig:related_works}(a)). 
Compared with the cascaded model that requires separate models for every different task, language model based approaches integrate diverse speech tasks into a generative language model, and potentially tackle different speech tasks with a single model, including speech synthesis, voice conversion, noise suppression, and speech editing. 
However, these speech language models utilize task-dependent prompt tokens for various speech tasks. This design ignores the inherent interrelationships among diverse speech tasks, which have the potential to significantly enhance the performance of each individual task. 
Take the voice conversion task as an example, the goal is to retain the linguistic information in speech while transforming non-linguistic elements. This requires the model to have capabilities like ASR models, where linguistic information is extracted, as well as abilities similar to TTS models, where speech is generated for a target speaker. Some of the previous voice conversion models are built using cascaded ASR and TTS model~\cite{VCC2020Huang,vcagents2021silvan,percascaded2022liao}. These previous approaches, with their separate handling of tasks, do not combine these overlapping skills, limiting the overall performance and efficiency of the model in tasks like voice conversion.
Considering this limitation, some works such as 
VoxtLM~\cite{Voxtlm2023Soumi} takes advantage of shared prompt tokens and achieves good performance in limited tasks.

In this work, we further exploit the capabilities of large language models across a wider range of speech tasks, and facilitate the straightforward integration of more tasks.
As shown in Figure~\ref{fig:related_works}(b), we introduce SpeechComposer, a speech language model that unifies common speech tasks by composing a fixed set of primary tasks.
We consider that speech tasks can be seen as composites of four primary speech tasks: speech language modeling (SpeechLM), text language modeling (TextLM), speech synthesis (TTS), and speech recognition (ASR). By composing the fixed set of uniform prompt tokens, the model can share knowledge between both primary tasks and composite tasks. Guided by different enrollment speech, SpeechComposer can be extended to many speech tasks, such as voice conversion (VC), speech enhancement (SE), speech editing, speech translation, and expressive speech synthesis. 
In this work, we particularly focus on VC and SE to demonstrate the effectiveness and scalability of SpeechComposer in handling primary tasks and composite tasks. VC is defined as a composition of ASR and TTS. Guided by an enrollment speech of the target speaker, SpeechComposer converts speech of the source speaker into those of the target speaker
through ASR and TTS tasks. SE is defined as the same composition, but with clean speech as the enrollment. 
The main contributions of this work are summarized as follows:
\begin{itemize}[nolistsep]
    \item We propose SpeechComposer, a unified language model that unifies common speech tasks and can be easily extended to more tasks through task compositions. SpeechComposer can handle composite tasks without changing the model structure and adding additional prompt tokens for specific tasks.
    \item  When expanding to new tasks, SpeechComposer leverages the capabilities of primary tasks through the composition of prompts, allowing for mutual enhancement between the primary and composite tasks. 
    \item To ensure reproducibility, we use publicly available datasets and an open-source toolkit for training and inference.
    The experimental results show that SpeechComposer performs well both in primary tasks and composite tasks. Also, SpeechComposer is flexible enough to be extended to new tasks in zero-shot scenarios.  Audio samples are 
available at demo page~\footnote{https://speechcomposer.github.io/samples/}.
\end{itemize}

\begin{figure*}[t]
    \centering
    \includegraphics[width=0.8\linewidth]{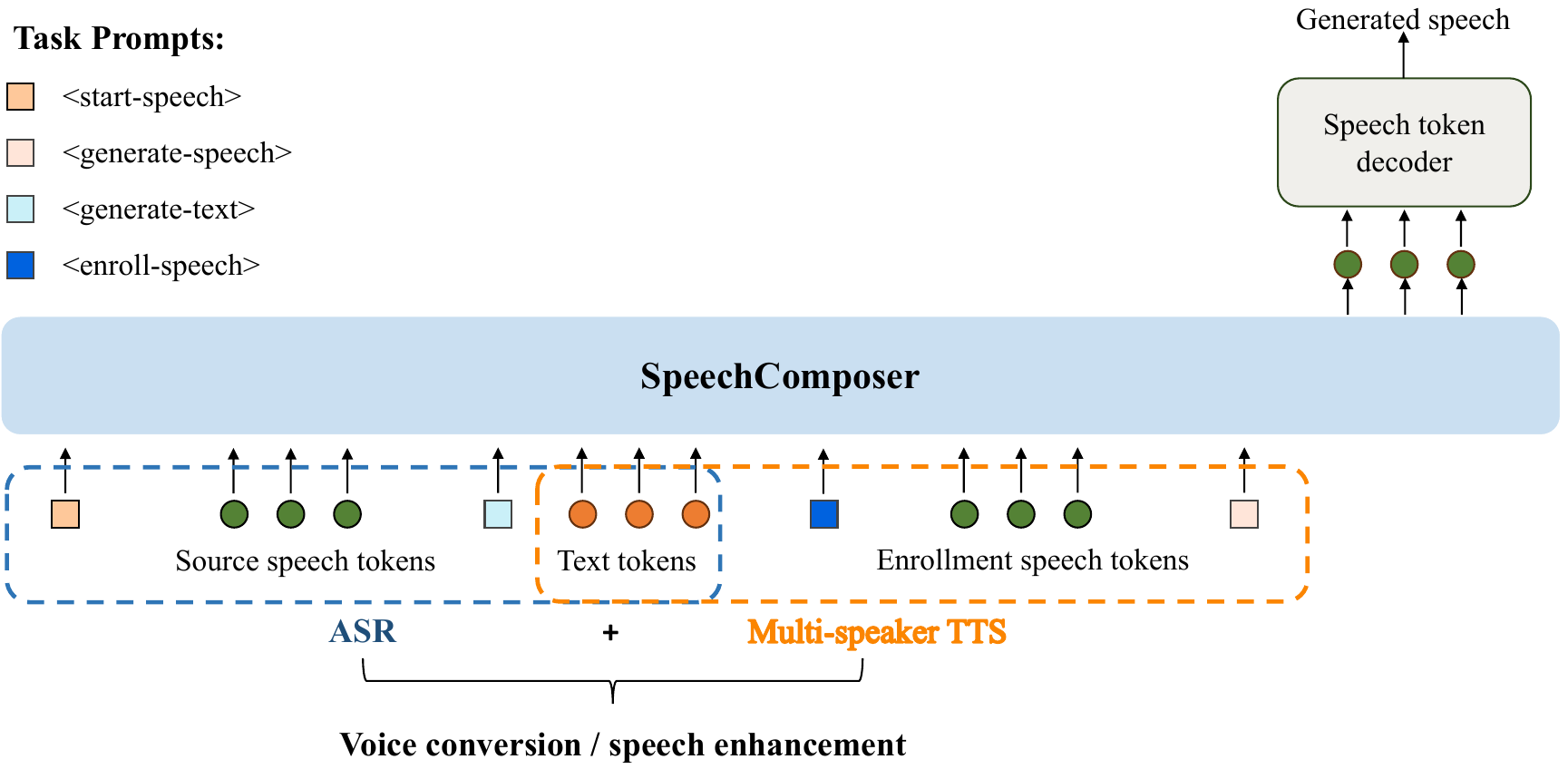}

    \caption{The overall architecture of SpeechComposer. In this picture, we use the task of voice conversion and speech enhancement as an example. It demonstrates how composite tasks can be accomplished through the composition of primary tasks and the use of prompt tokens. Speech enhancement can also be composed in a similar manner with different enrollment speech tokens.}
    \label{fig:model_arch}
\end{figure*}

\section{Related Work}
\subsection{Speech Language Models}
Transformer-based autoregressive language models (LM) show remarkable capacity in speech processing tasks. AudioLM~\cite{Audiolm2023Zal} explores the audio generation task as a language modeling task. Other works~\cite{agostinelli2023musiclm,Valle2023wang} further use similar autoregressive architecture in other audio related tasks, including music generation and zero-shot TTS. 
Furthermore, some works explore model performance in multi-task scenarios~\cite{makeavoice2023huang,polyvoice2023dong,SpeechX2023wang,Vallex2023zhang}. 
Make-a-Voice~\cite{makeavoice2023huang} explores both speech synthesis and voice conversion tasks in speech language modeling. SpeechX~\cite{SpeechX2023wang} explores speech synthesis, noise suppression, and speech editing tasks by designing task-dependent prompts. VioLA~\cite{Viola2023wang} and VALLE-X~\cite{Vallex2023zhang} integrate task IDs or language IDs to further perform speech translation tasks. These multi-task speech language models employ task-specific prompt tokens or patterns for different tasks, which ignores the inherent connections between these training tasks. Furthermore, for certain extended speech tasks like voice conversion, where paired training data is scarce, training a language model becomes even more challenging. In our work, instead of regarding these tasks as isolated components, we view speech tasks as combinations of primary tasks. This approach allows us to leverage the knowledge from primary tasks with abundant training data, facilitating easier task transfer and achieving better results.

\subsection{Composite Speech Task Frameworks}
Many previous studies employ cascaded models to accomplish speech related tasks. For example, some works link speech recognition and text-to-speech for speech enhancement and voice conversion tasks~\cite{VCC2020Huang,vcagents2021silvan,percascaded2022liao,MIMO_Speech-Chang2019}. For speech translation, previous works~\cite{integration2005Matusov} combine ASR with machine translation. Additionally, integrating ASR, machine translation, and TTS has been used for advanced speech translation tasks~\cite{prosodys2s2006Daniel}. However, these approaches usually rely on separate models for various sub-tasks and optimize them in stages independently. Compared with previous works, our method employs a unified framework to accomplish various tasks, allowing for mutual enhancement among cascaded tasks during the training process.

Also, some prior studies focus on training models for dual tasks or utilizing auxiliary tasks to enhance the performance of primary tasks. Chen et al.~\shortcite{se2015chen} explore the relationship between speech enhancement and automatic speech recognition. Ren et al.~\shortcite{unsupervisedTTS2019ren} and Hori et al.~\shortcite{cycleasr2019Hori} investigate the interplay between ASR and TTS. Zhang et al.~\shortcite{improving2019zhang} reveal that voice conversion performance could be enhanced through text supervision. Furthermore, other works link voice conversion with TTS~\cite{joint2019zhang}.
However, these studies generally design models for specific speech processing tasks, limiting their applicability to a broader range of tasks. Compared with these works, our proposed model alters the composition of tasks by changing the format of the input data, without redesigning the model for specific tasks.

\section{Proposed Method}
\subsection{Data Format}
\label{sec:data_format}

In the proposed SpeechComposer, various speech tasks are formulated as a unified language modeling task. To unify all tasks and corresponding data, we convert all speech signals into discrete tokens, allowing them to be modeled in a similar manner to the tokens in text processing. Suppose $Y=(y_{i} \in V_{\text{txt}} | i=1,...,t_{\text{txt}} )$ is a text sequence from a vocabulary $V_{\text{txt}}$, and $D=(d_{i} \in V_{\text{dst}} | i=1,...,t_{\text{dst}} )$ is a discrete speech token sequence from a vocabulary $V_{\text{dst}}$. We merge text and speech in a unified vocabulary $V_{\text{Comp}} = V_{\text{txt}} \cup V_{\text{dst}}$, and model them in the same manner. Then, we can model the probability of any sequence of speech or text tokens $Z = (z_{i} \in V | i=1,...,t)$ as $p(Z) = \prod_{i=1}^{t} p(z_{i}|z_{1},... z_{i-1})$,
where $Z$ can be discrete speech tokens $D$, text
tokens $Y$ or a combination of $Y$ and $D$.

To learn the interrelationship between different tasks, we incorporate new tasks through the method of task composition.
Being different from previous works~\cite{Viola2023wang,SpeechX2023wang,Vallex2023zhang}, we replace task-specific prompt tokens with composite ones based on a fixed set of prompts for different tasks. First, as shown in the upper part of Table~\ref{tab:data_format}, we identify four primary tasks: text language modeling (TextLM), speech-language modeling (SpeechLM), speech recognition (ASR), and speech synthesis (TTS).
Corresponding to these primary tasks, the following five primary prompt tokens are added to the unified vocabulary $V_{\text{Comp}}$ to distinguish various types of information. 
\begin{itemize}
    \item \texttt{\textlangle start-text\textrangle} and \texttt{\textlangle start-speech\textrangle} represent the start of a text or discrete speech sequence.
    \item \texttt{\textlangle generate-text\textrangle} and \texttt{\textlangle generate-speech\textrangle} indicate the data modality to be generated by the model. 
    \item \texttt{\textlangle enroll-speech\textrangle} indicates enrollment speech to provide context information for future token generation. Here, the goal of the enrollment speech is to provide additional context information, guiding it to generate the desired tokens for different tasks. More specifically, we first employ the speech tokenizer to encode continuous enrollment speech into discrete tokens. These tokens are then prefixed with the \texttt{\textlangle enroll-speech\textrangle} to signify the type of the information that follows.
\end{itemize}
Based on these primary tasks and prompt tokens, we define composite tasks by composing them, as shown in the lower part of Table~\ref{tab:data_format}. 
In this work, we expand four primary tasks to voice conversion and speech enhancement. 
However, it should be emphasized that SpeechComposer is flexible enough to be expanded to other speech tasks like speech translation, speech editing, and speech style transfer. 
Table~\ref{tab:data_format} shows that voice conversion is defined as a composition of ASR and TTS. Guided by the enrollment speech of the target speaker with \texttt{\textlangle enroll-speech\textrangle}, SpeechComposer converts discrete speech tokens of the source speaker into those of the target speaker through ASR and TTS tasks. Similarly, the task of speech enhancement operates on the same principle. Using clean audio as the enrollment speech for guidance, SpeechComposer utilizes ASR and TTS to generate clean, enhanced speech. It is important to note that for different tasks, the enrollment speech has a specific function depending on the task, but we always use the same \texttt{\textlangle enroll-speech\textrangle} token to maintain a consistent data format. We further compare SpeechComposer with other baseline models to show the efficacy of our designed data format in Section~\ref{sec:exp} related to RQ1.

\begin{table}[t]
\centering
\resizebox{\linewidth}{!}{\begin{tabular}{l|ccccc}
\toprule
\textbf{Model} & \textbf{Pretrained} & \textbf{Layers} & \textbf{Hidden size} & \textbf{Heads} & \textbf{Params} \\
\midrule
SpeechComposer-Base & \XSolidBrush & 12 & 768 & 12 & 125M \\
SpeechComposer-Pretrained & \Checkmark & 12 & 768 & 12 & 125M \\
SpeechComposer-Large & \Checkmark & 24 & 1024 & 16& 350M \\
\bottomrule
\end{tabular}
}
\caption{Details of SpeechComposer model variants.}
\label{tab:model_params}
\end{table}

\subsection{Model architectures}
A unified framework is designed to provide architecture compatibility across different tasks with a uniform data format. As shown in Figure~\ref{fig:model_arch}, SpeechComposer consists of three main components: speech tokenizer, decoder-only
language model and speech token decoder. Under this architecture, LLM can perceive both text and speech inputs and generates desired outputs for multiple tasks.

\paragraph{SpeechComposer.}
SpeechComposer consists of an embedding layer and a series of transformer~\cite{Tansformer2017Vaswani} decoder layers. First, the embedding layer $E$ maps the discrete token index sequence $Z$ (as shown in Section~\ref{sec:data_format}) into an $F$-dimensional feature space. As shown in Table~\ref{tab:model_params}, we set different $L$ transformer decoder layers and $H$ attention heads depending on the variants of SpeechComposer. 

\paragraph{Speech tokenizer.}
We use the speech tokenizer to convert continuous speech to discrete tokens, so that we can model them in a same manner with text tokens. Therefore, it is feasible to
perform speech tasks using the discrete speech tokens as input or with the corresponding text. We will discuss the effectiveness of extra text input in Section~\ref{sec:ablation}.
\paragraph{Speech token decoder.}
As shown in Figure~\ref{fig:model_arch}, we employ a pre-trained discrete token based vocoder to convert predicted discrete tokens to continuous speech. Here, we use the HiFiGAN~\cite{hifigan2020Kong} as the architecture and use x-vector~\cite{xvector2018snyder} as the speaker information.

\subsection{Training and Inference}
\label{sec:train&infer}
The model predicts a probability distribution over the tokens in $V_{\text{Comp}}$ by using a linear layer followed by softmax. Overall, SpeechComposer is trained as a language model in an autoregressive manner. Based on the data format defined in Section~\ref{sec:data_format}, we train our model on both primary tasks and composite tasks.
Specifically, for a composite task composed of $N$ primary tasks, its output is determined by the outputs of these $N$ subtasks autoregressively. Therefore, it can be formulated as:
\begin{equation}
\begin{split}
& p(Z^{\text{comp}}|Z^{\text{cond}},Z^{\text{enroll}}) \\ & = \prod_{n=1}^N p(Z^{\text{task}_n} | Z^{\text{task}_1}, \cdots, 
 Z^{\text{task}_{n-1}}; Z^{\text{cond}}; Z^{\text{enroll}}),
\end{split}
\end{equation}
where $Z^{\text{comp}}$ refers to the final prediction for the composite task, $Z^{\text{task}_1},..., Z^{\text{task}_{n-1}}$ are the outputs for previous $n-1$ primary tasks. Besides, $Z^{\text{cond}}$ refers to the condition information, which is the initial input, and $Z^{\text{enroll}}$ refers to the enrollment sequence, which can be both speech and text. Specifically, for voice conversion, $Z^{\text{enroll}}$ is any given speech of the target speaker represented as discrete tokens. For speech enhancement tasks, $Z^{\text{enroll}}$ is the discrete tokens of the corresponding speaker's clean speech.

In the training stage, we integrate $N$ primary tasks into a unified sequence. This composite method traditionally computes losses for all predicted components, which has been observed to induce training instability. To counteract this, we introduce a \textbf{randomized sampling strategy} for loss computation. The SpeechComposer generates predictions for the $N$ primary tasks, represented as $Z_{\text{pred}}^{\text{task}_{1}}, ..., Z_{\text{pred}}^{\text{task}_{N}}$, and the prediction for the composite task is denoted as $Z_{\text{pred}}^{\text{comp}}$. In essence, we allocate a series of hyper-parameters $q_1, q_2,\cdots, q_N, q_{\text{global}}$ to randomly select tasks for loss computation. Each $q_j$ corresponds to the likelihood that the loss for the $j$-th predicted part ($j\in\{1, \cdots, N\}$) or the global sequence ($j=\text{global}$) will be computed in a given training step, with the constraint that the sum of all $q_j$ values equals $1$. This ensures a probabilistically balanced loss calculation across all task predictions and the overall sequence, allowing for a more stable training process by preventing disproportionate weighting on any single task’s loss. We further design experiments to verify the effectiveness of the randomized sampling strategy in Section~\ref{sec:ablation}.

In the inference stage, we use beam search to predict discrete tokens in an autoregressive manner. Prediction of the desired task can be expressed as:
\begin{equation}
Z_{\text{pred}}^{\text{comp}} \leftarrow p(\cdot|Z^{\text{cond}}, Z_{\text{pred}}^{\text{task}_{1}}, ..., Z_{\text{pred}}^{\text{task}_{N-1}}, Z^{\text{enroll}})
\end{equation}
where $Z_{\text{pred}}^{\text{comp}}$ is the prediction of the desired task and 
$Z_{\text{pred}}^{\text{task}_{j}}$ refers to the prediction of the $j$-th primary task that forms part of the composite overall task.
For example, as shown in Table~\ref{tab:data_format}, for ASR, the condition is speech tokens $D$, while the prediction is the recognized text utterance $Y$. For TTS, the condition is the text utterance $Y$ and speech tokens from the target speaker as enrollment $D^{\text{enroll}}$, and the discrete tokens $D^{\text{tgt}}$ corresponding to the text utterance are predicted. 
As for composite tasks like speech enhancement, it is composed of two primary tasks, i.e. $N=2$. 
The model first predicts text $Y = Z_{\text{pred}}^{\text{task}_1}$ based on the discrete tokens of the noisy speech $D^{\text{noise}}$ and then uses the original condition information, the predicted text $Z_{\text{pred}}^{1}$, along with the enrollment speech $D^{\text{enroll}}$
as conditions to generate clean speech $D^{\text{clean}} = Z^{\text{comp}}$.

We train SpeechComposer using four primary tasks and two composite tasks, taking advantages of the interrelationships between primary tasks and composite tasks. 
In Section~\ref{sec:exp}, corresponding to RQ2, we design experiments with a variety of training task numbers in Section~\ref{sec:rq2}. This allows us to compare how different combinations of primary and composite tasks during training impact the task performance at the inference stage.
Additionally, we investigate the performance of SpeechComposer in a zero-shot scenario. 
In this context, zero-shot is defined as training SpeechComposer on primary tasks and a subset of composite tasks, and then testing it on other composite tasks that are unseen during training. Since no extra prompt tokens are added when expanding to new composites, we find SpeechComposer is sufficiently flexible to extend to unseen composite tasks. We will further discuss experimental results in Section~\ref{sec:rq3} corresponding to RQ3.

\section{Experiment Setup}

\subsection{Model configurations}
To train the sub-word model, we use all paired
text-speech datasets. For the speech tokenizer, we use the pretrained English HuBERT-Base model~\cite{hsu2021hubert} and use $k$-means clustering to discretize the 6$th$-layer embeddings. Here, we set $k=1000$ for $k$-means. Also, we convert the sampling rate of all speech data to 16kHz. 
As shown in Table~\ref{tab:model_params}, we train three variants of SpeechComposer with distinct configurations: SpeechComposer-base, SpeechComposer-pretrained, and SpeechComposer-large. Prior research~\cite{Textually2023Hassid} indicates that initializing a SpeechLM with a pretrained TextLM leads to enhanced performance and quicker convergence. Inspired by this methodology, for both SpeechComposer-pretrained and SpeechComposer-large, we initialize the SpeechComposer weights using the pretrained TextLM OPT~\cite{OPT} and construct the embedding table from the ground up.
We use 4 A100 GPUs for training small/medium models with the Adam optimizer~\cite{Adam2015Kingma} and a warmup learning rate scheduler.

\begin{table*}[t]
\centering
\resizebox{\linewidth}{!}
{
    \begin{tabular}{lcccccccccc}
    \toprule
    \multirow{2}{*}{\textbf{Setting}} & \multirow{2}{*}{\textbf{$\#$ params}} & \textbf{TextLM} & \textbf{SpeechLM} & \textbf{ASR} & \multicolumn{2}{c}{\textbf{TTS}} & \multicolumn{2}{c}{\textbf{Voice conversion}} & \multicolumn{2}{c}{\textbf{Speech enhancement}}\\
    & & PPL ($\downarrow$)  & PPL ($\downarrow$) & WER ($\downarrow$) & CER ($\downarrow$) & MOSNet ($\uparrow$) & CER ($\downarrow$) & MOSNet ($\uparrow$) & WER ($\downarrow$) & DNSMOS ($\uparrow$) \\
    \midrule
    \multirow{2}{*}{\textbf{Expert models}} & \multirow{2}{*}{---} & \multirow{2}{*}{---} & \multirow{2}{*}{---} & \textbf{dst-ASR-Hubert} & \multicolumn{2}{c}{\textbf{VITS}} & \multicolumn{2}{c}{\textbf{PPG-VC}} & \multicolumn{2}{c}{\textbf{USES}}  \\
    & & & & 4.2\slash 10.8 & 7.7 & 4.20 & 6.9 & 4.29 & 4.2 & 3.10 \\
    \midrule
    \textbf{VoxtLM-Base}$^\star$ & 125M & 18.3 & 36.7 & 4.7\slash 11.9 & 3.9 & 4.28 & --- & --- & --- & --- \\
    \textbf{VoxtLM-Large}$^\star$ & 350M & 16.4 & 32.3 & 3.6\slash 8.6 & 6.1 & 4.27 &--- & --- & --- & --- \\
    \midrule
    \midrule
    \textbf{SpeechComposer-Base} & 125M & 18.7 & 35.6 & 7.4\slash 15.7 & 3.2$^\dagger$ & 4.23 & 7.0 & 4.40$^\dagger$ & 8.3 & 3.08 \\
    \textbf{SpeechComposer-Pretrained}$^\star$ & 125M & 17.6$^\dagger$ & 34.7$^\dagger$ & 5.1\slash 13.0& 3.1$^\dagger$ & 4.24 & 7.8  & 4.36$^\dagger$ & 7.6 & 3.07 \\
    \textbf{SpeechComposer-Large}$^\star$ & 350M & 16.8 & 34.0 & 6.4\slash 16.9 & 3.4$^\dagger$ & 4.26 & 6.7$^\dagger$ & 4.35$^\dagger$ & 7.6 & 3.08 \\
    \bottomrule
    \end{tabular}
}
\caption{Experimental results comparing SpeechComposer with language model based baselines for different tasks (primary tasks and composite tasks). For ASR, we report test-clean/test-other results. Here, $^\star$ denotes initialization with OPT, and $^\dagger$ signifies that SpeechComposer achieves better results compared with the VoxtLM baselines that have the same model sizes or the expert models.}
\label{tab:main_results}
\end{table*}

\subsection{Datasets}
We use a combination of speech-only, text-only, paired speech-text, and paired speech-text datasets for SpeechComposer's training and inference. All datasets we used are public corpora.
\paragraph{SpeechLM.} We use speech-only dataset LibriLight~\cite{Librilight2020} for speech language modeling task. Librilight contains more than 60K hours of speech data from 7,439 speakers.
\paragraph{TextLM.} We use the text portion of LibriSpeech~\cite{Librispeech2015} dataset as the training data for a text language modeling task. It contains about 40M text utterances.
\paragraph{Speech recognition.}
We use paired speech-text dataset Librispeech~\cite{Librispeech2015} for a speech recognition task. It contains 960 hours of speech data with 281K utterances.
\paragraph{Multi-speaker TTS.}
For speech synthesis, we use two multispeaker datasets,  LibriTTS~\cite{LibriTTS2019} and Hi-Fi TTS. LibriTTS is a multi-speaker English corpus of approximately 585 hours of read English speech from 2456 speakers. Hi-Fi TTS contains about 292 hours of speech from 10 speakers with at least 17 hours per speaker.
\paragraph{Voice conversion.}
For voice conversion, we use parallel datasets which have pairs of audio recordings where the same linguistic content is spoken by different speakers. We use VCTK~\cite{veaux2017cstr}, CMU-Arctic~\cite{cmuarctic2004kominek}, VCC 2018~\cite{vcc18}, VCC 2020~\cite{vcc2020}. 
The VCTK dataset contains 44 hours of studio recorded data from 109 speakers. These speakers read the same Rainbow Passage and elicitation paragraph. CMU-Arctic contains about 3 hours of speech from three speakers. VCC 2018 and VCC 2020 come from Voice conversion challenges 2018 and 2020 respectively.
We obtain about 50 hours of parallel speech data in total.
\paragraph{Speech enhancement}
For the speech enhancement task, we use Voice Bank+DEMAND dataset~\cite{VCTKnoisy2016}, which is a typical SE database with clean and noisy parallel speech. It contains about 8.8 hours of speech data from 28 speakers. 
There are around 400 sentences available from each speaker.

\section{Evaluation metrics}
To ensure consistency and reproducibility, we employ objective metrics for each task. 
\paragraph{Word Error Rate (WER)\slash Character Error Rate (CER)}
We use WER and CER to assess the performance of the ASR model and the
intelligibility of the generated speech to the given transcription. 
For the evaluation of generated speech, we use Whisper Large v2\footnote{https://huggingface.co/openai/whisper-large-v2}~\cite{whisper2023Radford} model to transcribe the generated speech and calculate the WER score.

\paragraph{Speaker similarity score.}
The speaker similarity score is used to evaluate how well the generated speech matched the specific characteristics of the speaker. This score is derived from the cosine similarity between the speaker embeddings of both the generated and the target speech signals. Here, we calculate the x-vector as the speaker embedding. The speaker similarity score (Sim-\textit{speech}) metric is applied in tasks including multi-speaker TTS and voice conversion. Besides, we use the TER (token error rate) to evaluate the similarity between the predicted discrete speech tokens and the actual discrete speech tokens.

\paragraph{MOSNet.}
Following ~\cite{Voxtlm2023Soumi}, we use automatic mean opinion score MOSNet~\cite{Mosnet2019Lo,MOSpred2022Cooper} to evaluate the quality of generated speech.
\paragraph{DNSMOS.}
Following previous works, we use DNSMOS (OVRL)~\cite{dnsmos2022Reddy} 
to evaluate the model's performance in the speech enhancement task. DNSMOS is a non-intrusive perceptual objective metric, which is used to simulate the human subjective evaluation on the DNS blind test set. More specifically, we leverage the OVRL score from the DNSMOS P.835 model\footnote{https://github.com/microsoft/DNS-Challenge/}.

\paragraph{Perplexity.}
Perplexity is a widely used metric in natural language processing, especially for evaluating language models. It measures how well a language model predicts a sample.  
We use perplexity to evaluate the performance of speech language modeling and text language modeling.

\subsection{Evaluation metrics}
To ensure consistency and reproducibility, we employ objective metrics for each task, as shown in Table~\ref{tab:data_format}. 
\paragraph{Word Error Rate (WER) \slash Character Error Rate (CER)}
We use WER and CER to assess the performance of the ASR model and the
intelligibility of the generated speech to the given transcription. 
For the evaluation of generated speech, we use Whisper Large v2\footnote{https://huggingface.co/openai/whisper-large-v2}~\cite{whisper2023Radford} model to transcribe the generated speech.

\paragraph{Speaker similarity score.}
Speaker similarity scores measure the match of generated speech to the speaker's characteristics. This score is derived from the cosine similarity between the speaker embeddings of both the generated and the target speech. Here, we calculate the x-vector as the speaker embedding. The speaker similarity score (Sim-\textit{speech}) is applied in multi-speaker TTS and VC. Besides, we use the TER (token error rate) to evaluate the similarity between the predicted discrete speech tokens and the actual discrete speech tokens.

\paragraph{MOSNet.}
Following ~\cite{Voxtlm2023Soumi}, we use automatic mean opinion score MOSNet~\cite{Mosnet2019Lo} to evaluate the quality of the generated speech in TTS and VC tasks.
\paragraph{DNSMOS.}
Following previous works, we use DNSMOS (OVRL)~\cite{dnsmos2022Reddy} 
to evaluate the model's performance in SE tasks. DNSMOS is a non-intrusive perceptual objective metric, which is used to simulate the human subjective evaluation on DNS blind test sets. More specifically, we leverage the OVRL score from the DNSMOS P.835 model\footnote{https://github.com/microsoft/DNS-Challenge/}.

\paragraph{Perplexity.}
Perplexity is a widely used metric in natural language processing, especially for evaluating language models. It measures how well a language model predicts a sample. 
We use perplexity to evaluate the performance of speech language modeling and text language modeling.

\section{Experiment results}
\label{sec:exp}
We focus on the following research questions:

\textbf{RQ1}: With the unified data format, can SpeechComposer achieve better performance for both primary tasks and composite tasks?

\textbf{RQ2}: Does the number of tasks trained in the model and the quantity of training data have an impact on the performance of primary and composite tasks?

\textbf{RQ3}: Is the task composition design in SpeechComposer robust enough to generalize to new tasks in a zero-shot manner?

\subsection{Performance on primary tasks (RQ1)}

Table~\ref{tab:main_results} shows the performance of SpeechComposer in various tasks compared to the baseline models, including both speech language models\footnote{Due to the lack of open-source availability for related works such as SpeechX, VioLA, and Make-a-Voice, we are unable to conduct a fair comparison with them. Therefore, in this context, we mainly compare our work with the open-source project VoxtLM (https://github.com/espnet/espnet/pull/5472).} and individual expert models. SpeechComposer achieves comparable or even better performance than speech language model baselines that have the same model sizes or the expert models, especially in the TTS task. Based on the performance of different models in primary tasks as presented in Table~\ref{tab:main_results}, we have the following observations:
\paragraph{Comparison with expert models.} In TTS tasks, we compare SpeechComposer with VITS~\cite{VITS2021Kim}, a parallel end-to-end TTS model that can generate high quality natural speech. we create a test set of 270 utterances from two speakers from the
LibriTTS test-clean\footnote{Specifically, speaker ids are 1089 and 1284.}. Compared to VITS, SpeechComposer gets a superior MOSNet score that shows higher speech quality. It is noteworthy that SpeechComposer-Pretrained significantly outperforms in terms of intelligibility, evidenced by a substantial reduction in the Character Error Rate (CER) from 7.7\% to 3.1\%. Though SpeechComposer is trained with a larger dataset compared to VITS and YourTTS, it should be mentioned that for TTS models, having diverse training data with more noise and more speakers often degrades rather than improves the performance~\cite{Voxtlm2023Soumi}. As for the ASR tasks, we use the same expert model with VoxtLM, dst-ASR-Hubert~\cite{dsthubert2023Chang}. Comparing the results of SpeechComposer with dst-ASR-Hubert, SpeechComposer does not achieve better performance in ASR tasks. We assume this might be related to the training method of the language model, making it more suited for generative tasks in speech. From previous works~\cite{Voxtlm2023Soumi}, we suggest that the performance in ASR tasks could potentially be improved by increasing the ASR training data.

\paragraph{Comparison with language based models.} 
We further compare SpeechComposer with the unified speech language model VoxtLM~\cite{Voxtlm2023Soumi}.
Under similar parameter conditions, SpeechComposer demonstrates much better performance in the TTS tasks. For example, SpeechComposer-Pretrained demonstrates better intelligibility in the TTS task, in terms of CER from 3.9\% to 3.1\%. Also, SpeechComposer-Large achieves significant improvement in intelligibility, in terms of CER from 6.1\% to 3.4\%
and comparable speech quality. Besides, SpeechComposer-Pretrained achieves better performance in TextLM and SpeechLM tasks. This demonstrates that for primary tasks, SpeechComposer benefits from a unified training approach, achieving better results compared to language model based baselines that have the same model sizes, particularly in generative tasks.

\paragraph{Comparison of model size}
Furthermore, we compare SpeechComposer with different parameters and training strategies. Comparing SpeechComposer-Base and SpeechComposer-Pretrained, we observe that initialization with OPT can improve the performance of TextLM, SpeechLM, ASR, and TTS. What's more, by comparing SpeechComposer-Large with SpeechComposer-Pretrained, we observe that a larger model size can improve the performance of TextLM, SpeechLM, and ASR.

\subsection{Performance on Composite Tasks (RQ1)}
To answer RQ1, we further compare SpeechComposer with language model based models and expert models on composite tasks, i.e., voice conversion and speech enhancement.
\paragraph{Comparison with expert models.} In the VC task, we compare SpeechComposer with an expert model PPC-VG~\cite{PGGVC2021Liu}, a voice conversion model based on phonetic posterior-gram (PPG). We split the test set from the VCTK corpus, which includes conversions from seen speakers to seen speakers. Each test set contains 350 paired speech samples.
The experimental results are shown in Table~\ref{tab:main_results}. Compared to the expert model, the SpeechComposer-base model achieves better speech quality, while SpeechComposer-Large shows better intelligibility in terms of CER from 6.9\% to 6.7\%. Especially in Sim-\textit{speech}, SpeechComposer-base outperforms PPG-VC from 0.63 to 0.74.
For speech enhancement, we compare with USES~\cite{zhang2023universal},  an unconstrained speech enhancement and separation network that achieves high performance across different conditions.
We use the test set of Voice Bank+DEMAND, which has 824 utterances of two speakers from the Voice Bank corpus mixed with unseen DEMAND noises. Compared to the USES,  SpeechComposer's effectiveness is slightly inferior to that of an expert model specifically designed for speech enhancement tasks.

\begin{table*}[t]
\centering
\resizebox{\linewidth}{!}
{
    \begin{tabular}{lccccccc}
    \toprule
    \multirow{2}{*}{\textbf{Training tasks}} & \multicolumn{2}{c}{\textbf{TTS}} & \multicolumn{3}{c}{\textbf{Voice conversion}} & \multicolumn{2}{c}{\textbf{Speech enhancement}}\\
     & MOSNet ($\uparrow$) & CER ($\downarrow$)  & MOSNet ($\uparrow$) & CER ($\downarrow$) & Sim-\textit{speech}($\uparrow$) &  DNSMOS ($\uparrow$) & WER ($\downarrow$)  \\
    \midrule

     TTS & 4.20 & 3.9 & --- & --- & --- & --- & --- \\
     VC & --- & --- &\multicolumn{3}{c}{------ Cannot converge -----} & --- & --- \\
     SE & --- & --- & --- & ---& --- & \multicolumn{2}{c}{------ Cannot converge -----} \\
     TTS + VC & 4.22 & 5.1 & 4.36 & 72.6 & 0.69 & --- & ---  \\
     TTS + SE & \textbf{4.25} & 3.4 & --- & --- & ---& \textbf{3.07} & 32.6\\
     TTS + ASR + VC  & 4.23 & 5.3 & 4.36 & 9.0 & \textbf{0.71} & --- & --- \\
     TTS + ASR + SE & 4.24 & 3.8 & --- & --- & --- & 3.04 & 28.2  \\
     TTS + ASR + VC + SE & 4.21 & \textbf{3.1} & \textbf{4.38} & \textbf{7.8} & 0.70 & \textbf{3.07} & \textbf{7.9} \\
     \midrule
     SpeechComposer-Base \textit{w/o} VC & --- & --- & 4.37 & 12.1 & 0.73 & --- & ---  \\
     SpeechComposer-Base \textit{w/o} SE & --- & --- & --- & --- & --- & 3.03 & 29.4 \\

    \bottomrule
    \end{tabular}
}

\caption{Experimental results of SpeechComposer with different training tasks. The bolded numbers highlight the best performance achieved for each metric. Generally, the more tasks a model is trained on, the better its performance.}

\label{tab:task_num}
\end{table*}

\begin{table}[t]
\centering
\resizebox{\linewidth}{!}
{
    \begin{tabular}{lccccc}
    \toprule
    \multirow{2}{*}{\textbf{Prompt}} & \multicolumn{3}{c}{\textbf{Voice conversion}} & \multicolumn{2}{c}{\textbf{Speech enhancement}}\\
     & MOSNet ($\uparrow$) & CER ($\downarrow$) & Sim-\textit{speech}($\uparrow$) &  DNSMOS ($\uparrow$) & WER ($\downarrow$)  \\
    \midrule

     w/ text & 4.35 & 1.1 & 0.72 & 3.11 & 3.4  \\
     w/o text & 4.32 & 7.0 & 0.71 & 3.09 & 8.1 \\
    \bottomrule
    \end{tabular}
}

\caption{Results on voice conversion and speech enhancement with or without text input. It shows that leveraging the text input is particularly beneficial for enhancing the intelligibility of the output speech. }

\label{tab:text_prompt}
\end{table}

\begin{table}[t]
\centering
\resizebox{\linewidth}{!}{\begin{tabular}{lcccc}
\toprule
 \textbf{Settings} & \textbf{MOSNet} ($\uparrow$) & \textbf{CER} ($\downarrow$) & \textbf{Sim-\textit{speech}} ($\uparrow$) & \textbf{TER} ($\downarrow$) \\
\midrule
seen speakers & 4.35\slash 4.32 & 1.1\slash 7.0 & 0.72\slash 0.71 & 0.63\slash 0.70 \\
unseen speakers & 4.36\slash 4.28 & 1.5\slash 9.0 & 0.68\slash 0.65 & 0.69\slash 0.72    \\
\bottomrule
\end{tabular}
}

\caption{Results of voice conversion between seen speaker and unseen speakers, including both speech-text\slash speech-only results.}

\label{tab:vc_speakers}
\end{table}

\begin{table}[t]
\centering
\resizebox{\linewidth}{!}{\begin{tabular}{lccccc}
\toprule
 \textbf{Settings} & Training tasks & \textbf{MOSNet} ($\uparrow$) & \textbf{CER} ($\downarrow$) & \textbf{Sim-\textit{speech}}($\uparrow$) & \textbf{TER} ($\downarrow$) \\
\midrule
Base & all tasks & 4.35 \slash 4.32 & 1.1 \slash 7.0 & 0.72 \slash 0.71 & 0.63 \slash 0.70 \\
$\quad$ w/o rs & all tasks  & 4.33 \slash 4.32 &1.7 \slash 7.3 & 0.71 \slash 0.70 & 0.65 \slash 0.67 \\
Base & VC & 4.38 \slash 4.37 & 25.3 \slash $-$ & 0.72 \slash $-$& 0.89\slash $-$ \\
$\quad$ w/o rs & VC & 4.35 \slash4.35 & 62.7 \slash $-$ & 0.70 \slash $-$ & 0.99 \slash $-$   \\
\bottomrule
\end{tabular}
}

\caption{Effects of randomized sampling strategy with different amount of training data. Here, ``Base'' refers to SpeechComposer-Base, and ``rs'' is the abbreviation of randomized sampling strategy.}

\label{tab:loss}
\end{table}

\paragraph{Comparison of model size.} 
We further explore if a larger model size can help composite tasks by comparing SpeechComposer-Pretrained and SpeechComposer-Large. For both VC and SE, we observe that the larger model achieves better intelligibility and comparable speech quality. 

\subsection{The Effects of Training Tasks Numbers. (RQ2)}
\label{sec:rq2}
As mentioned in Section~\ref{sec:train&infer}, we also conduct experiments where we use different numbers of tasks during training to explore the potential interactions between different primary tasks and composite tasks.
Table~\ref{tab:task_num} shows the experimental results. We have the following observations.
(1) Increasing the number of training tasks and the corresponding volume of training data can enhance the performance of both primary tasks (TTS) and composite tasks (VC and SE). For instance, as the number of training tasks increases, the CER metric for the TTS task improves from 3.9\% to 3.1\%. For VC and SE tasks, we observe that training only on the VC or SE task with related datasets results in the model's inability to converge.
This is attributed to the insufficient training data compared to the large model size.  This further emphasizes the importance of training data volume in language model training. 
(2) Adding primary tasks can improve the performance of composite tasks. As shown in Table~\ref{tab:task_num}, introducing the TTS task adds more training data, which enhances the speech quality of the VC and SE tasks. Additionally, the ASR task significantly improves the intelligibility of VC and SE in terms of CER and WER metrics. It also highlights the necessity of utilizing primary tasks when expanding to new tasks.
(3) Adding composite tasks can also improve the performance of primary tasks. From Table~\ref{tab:task_num}, we can observe that introducing the speech enhancement task significantly enhances the quality and intelligibility of the generated speech, for both TTS and VC tasks. For example, 
when trained only with the TTS task, the CER is 3.9\%. However, after training with TTS and speech enhancement tasks combined, the CER improved to 3.4\%. Voice conversation tasks have similar results to the former.

\subsection{The Performance of Zero-shot Transfer (RQ3)}
\label{sec:rq3}
To answer RQ3, we experiment with zero-shot generation on tasks for which the model was not specifically trained. Here, we conduct evaluations on two composite tasks, SE and VC, comparing their performance with scenarios where training data is available.
As indicated in the last two rows of Table~\ref{tab:task_num}, SpeechComposer-Base is not trained on SE or VC tasks respectively, but it is tested on these two tasks. 
The results reveal that the model is capable of performing zero-shot SE and VC tasks due to our design of uniform prompt tokens, even without explicit training on such tasks. For both tasks, zero-shot generation results in a noticeable increase
in WER or CER, whereas the degradation in DNSMOS and MOSNet scores is modest. 
It is noteworthy that in the VC task, the models not trained on the VC task outperformed those trained solely on VC or VC+TTS tasks. This underscores the importance of data quantity for speech language models. Also, following this zero-shot manner, our model is flexible enough to extend to more composite tasks, even in the absence of task-specific training data.

\subsection{Ablation Studies}
\label{sec:ablation}
\paragraph{Effectiveness of additional text input.}
With the data format design of SpeechComposer described in Section~\ref{sec:data_format}, it is feasible to
perform VC and SE using the discrete speech tokens as input or with the corresponding text. To verify the efficacy of incorporating additional text input in the SpeechComposer, we conduct experiments on VC and SE tasks. We consider two scenarios -- using only speech input and using combined speech and ground truth text input. 
The experimental results are presented in Table~\ref{tab:text_prompt}. For both
tasks, omitting the text input resulted in a noticeable increase in CER or WER, whereas the degradation in MOSNet or DNSMOS scores is modest. These findings suggest that leveraging the text input is particularly beneficial for enhancing the intelligibility of the output speech. 

\paragraph{Performance of VC for unseen speakers.}
To validate the SpeechComposer's performance on VC tasks between unseen speakers, we split 350 paired speech samples between unseen speakers from the VCTK corpus as the test set.
As shown in Table~\ref{tab:vc_speakers}, we observe that SpeechComposer performs well in VC between unseen speakers. Compared with VC between seen speakers, VC between unseen speakers is more difficult, leading to a slightly lower Sim-\textit{speech} and higher TER metrics.
\paragraph{Effectiveness of randomized sampling strategy.}
We conduct experiments to verify the effectiveness of our proposed randomized sampling loss on the VC task. For the hyper-parameters $q_{1},\cdots, q_{N}$ and $q_{\text{global}}$ of randomized sampling strategy mentioned in Section~\ref{sec:train&infer}, we set $q_{1}=q_{2}=0.3$, $q_{\text{global}}=0.4$ respectively .
As shown in Table~\ref{tab:loss}, we observe that removing the randomized sampling loss will lead to worse intelligibility, especially when we use few training data. For example, removing the randomized sampling loss trained on the VC task leads to a noticeable increase in the CER score (from 25.3\% to 62.7\%). This demonstrates that the quantity of training data contributes to training stability and validates the effectiveness of the randomized sampling.

\section{Conclusion}
In this paper, we describe SpeechComposer, a decoder-only language model that unifies common speech tasks by composing a fixed set of prompt tokens. By defining four primary tasks, speechLM, textLM, TTS, and ASR, SpeechComposer can easily extend to more speech tasks, including VC and SE. 
These tasks can share the knowledge of each other, and achieve better performance.
We demonstrate SpeechComposer's efficacy through comprehensive experiments and analyze the impact of various training tasks on model performance. Also, we verify the behavior of composite tasks in zero-shot scenarios, which shows the flexibility of SpeechComposer to extend to more new tasks.
We will further explore this work by expanding supported tasks, taking advantage of the knowledge of pretrained foundation models and enhancing robustness.
\section{Acknowledgements}
Experimnets of this work used the Bridges2 system at PSC and Delta system at NCSA through allocations CIS210014 and IRI120008P from the Advanced Cyberinfrastructure Coordination Ecosystem: Services \& Support (ACCESS) program.
\newpage
\bibliographystyle{named}
\bibliography{ijcai24}
\end{document}